\def\year{2022}\relax
\documentclass[letterpaper]{article} 
\usepackage{aaai22}  
\usepackage{times}  
\usepackage{helvet}  
\usepackage{courier}  
\usepackage[hyphens]{url}  
\usepackage{graphicx} 
\usepackage{natbib}  
\usepackage{caption} 
\DeclareCaptionStyle{ruled}{labelfont=normalfont,labelsep=colon,strut=off} 
\usepackage{algorithm}
\usepackage{algorithmic}

\usepackage{newfloat}
\usepackage{listings}
\floatstyle{ruled}
\newfloat{listing}{tb}{lst}{}
\floatname{listing}{Listing}

\usepackage{authblk}
\usepackage{amsmath,amssymb}

\usepackage{multirow}
\usepackage{subcaption}
\usepackage{pifont}
\title{Low-rank Tensor Decomposition for Compression of Convolutional Neural Networks Using Funnel Regularization}
\author{
    Bo-Shiuan Chu,
    Che-Rung Lee
}
\affiliations{

}

\begin{document}

\maketitle

\begin{abstract}
Tensor decomposition is one of the fundamental technique for model compression of deep convolution neural networks owing to its ability to reveal the latent relations among complex structures. However, most existing methods compress the networks layer by layer, which cannot provide a satisfactory solution to achieve global optimization.  In this paper, we proposed a model reduction method to compress the pre-trained networks using low-rank tensor decomposition of the convolution layers.  Our method is based on the optimization techniques to select the proper ranks of decomposed network layers.  A new regularization method, called funnel function, is proposed to suppress the unimportant factors during the compression, so the proper ranks can be revealed much easier.  The experimental results show that our algorithm can reduce more model parameters than other tensor compression methods.  For ResNet18 with ImageNet2012, our reduced model can reach more than 2 times speed up in terms of GMAC with merely 0.7\% Top-1 accuracy drop, which outperforms most existing methods in both metrics.  
\end{abstract}

\section{Introduction}
Despite the impressive accuracy of deep convolution neural networks (CNNs), the computational cost and memory consumption of those CNNs have made them difficult to be applied on resource limited devices.  Methods including pruning and quantization have been widely studied to reduce the model sizes \cite{han2015deep, choi2016towards, xu2018deep, Deng2020survey}.  Later model compression methods tend to preserve the model structure for better hardware utilization \cite{luo2017thinet} \cite{he2017channel} \cite{zhuang2018discrimination}.  One of the structure preserving model compression methods is using low-rank approximation \cite{lebedev2014speeding, kim2015compression, tai2015convolutional, gusak2019automated}, which has the potential to reveal latent relations among underlying structures and to achieve a better compression ratio.  

However, exiting model compression method for convolution layers based on low-rank approximations face many computational challenges \cite{cheng2017survey, Deng2020survey}, which have hindered this direction.  First, exiting methods usually perform the low-rank approximation layer by layer, so it is hard to achieve global optimization for model compression.  When the number of layers increases, those methods suffer large accuracy loss and long processing time.  Second, the compressed model usually requires a lot of retraining to recover the original accuracy, which makes it hard to compress a large model.  Last, the rank of a high order tensor is not well defined \cite{kolda2009tensor}.  For the applications of model compression, the situation becomes more complex, because the elements of a network are changed over epochs. 



In this paper, we proposed a model reduction method to compress the pre-trained networks using low rank approximation for the convolution layers. The major differences distinguishing our work from the previous ones is that our method considers the global optimization of all layers during the model compression.  In addition, unlike existing methods which trim the network after tensor decomposition and re-train the model to retain the original accuracy, our method factorizes the network first, and instruments a regularization gate to each factor of the decomposed tensor.  After that, our method trains the factorized model to attain a higher accuracy than the original model.  The pruning process is based on the values of the regularization gates, which removes the factors with small gate values. Last, to avoid the difficulty of selecting a proper rank, we designed a new regularization function, called funnel function, which can better separate the desired and undesired factors. 

\begin{figure}[tb]
\centering
\includegraphics[scale = 0.5, clip]{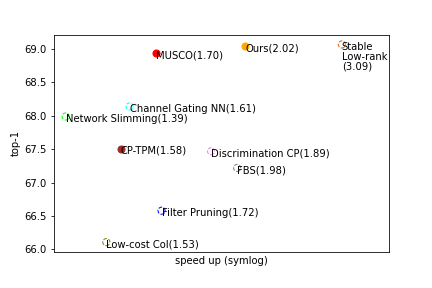}
\caption{Comparison between our method and state-of-the-art compression methods on ResNet18 using ImageNet dataset \cite{he2017channel}\cite{zhuang2018discrimination}\cite{gusak2019automated}\cite{liu2017learning}\cite{dong2017more}\cite{hua2019channel}\cite{gao2018dynamic}\cite{astrid2017cp}\cite{stable2020}. Our method provides 2.02x speed up in GMAC, defined in (\ref{eqn:speedup}),  and 0.72\% accuracy loss.}
\label{fig:compare}
\end{figure}

We have conducted experiments to evaluate the effectiveness of our method.  The results show that our method can reduce more model parameters than other tensor compression methods for various models. For ResNet18 with ImageNet dataset, our method  can achieve more than two times speed up in terms of GMAC with similar Top-1 accuracy.  Figure \ref{fig:compare} compares the results of our algorithm with other methods in terms of compression ratio (speed-up of GMAC) and Top-1 accuracy.  Our method outperforms most of the method except the stable low-rank method \cite{stable2020}, whose values are reported from their paper.

The rest of this paper is organized as follows.  Section \ref{chap:related} introduces related work for low-rank approximation and rank selection.  Section \ref{chap:method} illustrates our method.  Section \ref{chap:exp} presents the experimental results.  The conclusion and the future work are given in the last Section.

\section{Related Work}

\label{chap:related}
CNNs are mainly consisted of convolution layer and fully-connected layer, where convolution layers dominate most computation cost of inference. Misha Denil \cite{denil2013predicting} showed most model parameters can be predicted by a small subset of model parameters, which indicate the possibility of removing redundant model parameters.  Emily Denton \cite{denton2014exploiting} applied truncated singular value decomposition (SVD) on the weight matrix of fully-connected layer, which does not cause a significant drop in accuracy. After that, various methods were also proposed \cite{gong2014compressing} \cite{chen2015compressing} \cite{cheng2015fast} \cite{novikov2015tensorizing}, showing better compression capability than SVD.

Several methods based on low-rank decomposition of convolution kernel tensor were also considered to accelerate convolution operations. A tensor is defined as a multi-dimensional array. Canonical Polyadic Decomposition (CPD) \cite{carroll1970analysis} \cite{harshman1994parafac} \cite{shashua2005non} 
and Tucker Decomposition (TKD) \cite{tucker1966some} \cite{de2000multilinear} \cite{kim2007nonnegative} are two most popular tensor decomposition methods.  
Vadim Lebedev \cite{lebedev2014speeding} used Canonical Polyadic Decomposition (CPD) to speed up CNNs.  Due to the instability of CPD, only the result on single layer is reported.  This problem of CPD has been addressed in \cite{stable2020}, in which the first order perturbation is used to stabilize CPD. Yong-Deok Kim \cite{kim2015compression} used Tucker Decomposition (TKD) to speed up CNNs. The ranks of convolution kernels are determined by the solution of Variational Bayesian Matrix Factorization (VBMF) \cite{nakajima2012perfect} on unfolded tensor in different modes.  However, the rank determined by VBMF is too aggressive to recover the original accuracy.  In MUSCO \cite{gusak2019automated}, the authors proposed the concept of weakened rank and extreme rank, where extreme rank is determined by VBMF. The ranks of network are then gradually reduced from weakened rank to extreme rank.  In \cite{stable2020}, authors presented a stablization method for CP to achieve better compression ratio.  However, it still uses brute force method for rank selection.

In this paper, we employed TKD for model compression, since it generally performs better than CPD for data compression \cite{kolda2009tensor}.  As an extension of SVD, TKD computes the orthonormal basis of different modes of a tensor. The core tensor obtained in TKD can be seen as a compressed form of the original tensor.  However, TKD of a tensor is not unique.  There are various methods to compute TKD of a given tensor $T$ \cite{kroonenberg1980principal} \cite{kapteyn1986approach} \cite{de2000best} \cite{elden2009newton}. Here we adopt HOSVD \cite{de2000multilinear} to calculate the TKD of a tensor $T$, where the factor matrices of TKD are computed as the $R^n$ leading left singular vectors of the mode-n unfolded $T$. The value of $R^n$ is the approximation rank of mode-n unfolded $T$ \cite{kolda2009tensor}.  When $R^n$ is smaller than the rank of mode-n unfolded $T$ for at least one mode, the decomposition is called truncated HOSVD, which is not optimal in terms of the norm difference.  

Other tensor decomposition methods, such as Block tensor decomposition \cite{blockTKD} and Tensor Train decomposition \cite{TT2009}, are not considered.  Block tensor decomposition requires to select a proper block size, which may complicated the problem; and the Tensor Train decomposition is more suitable higher order tensors.  For CNNs, the orders of tensors are usually less than or equal to four.

\section{Method}
\label{chap:method}
Conventional model compression methods based on low-rank approximation usually utilize the iterative approach to reduce the model size \cite{gusak2019automated}.  First they decompose one or some layers and prune the factors with lower rank.  Then, they retrain the model to recover the accuracy.  Such processes are performed iteratively until convergence.  The iterative methods have several drawbacks.  First, the pruning is based on rank of the original model, whose number of parameters and their values would be changed after decomposition and retraining.  Second, the training process that performs low-rank approximation layer by layer overlooks the interconnection among different layers, and therefore is hard to achieve global optimization.    

To overcome those difficulties, we proposed a new pruning flow.  Given a pre-trained model as the input, our method consists of four steps:
\begin{enumerate} 
    \item Decomposition: decomposes each layer of the model with full rank.
    \item Training:  trains the decomposed model to obtain a higher accuracy.
    \item Compression: applies regularization technique to measure the importance of each decomposed factor, and removes the redundant ones for compression.
    \item Fine-tuning: fine tunes the whole model to achieve higher accuracy and lower computation cost.
\end{enumerate}
Below we introduce the details of each step.

\subsection{Decomposition}
The proposed training flow can be applied to various tensor decomposition methods. Here we used Tuker decomposition (TKD) to illustrate the procedure.  The convolution operation in CNNs is a linear mapping from tensor $X \in R^H{}^\times{}^W{}^\times{}^S{}$ to tensor $Y \in R^H{}^\times{}^W{}^\times{}^T{}$ with a kernel tensor $K \in R^D{}^\times{}^D{}^\times{}^S{}^\times{}^T{}$,
\begin{equation}
Y_{h,w,t} = \sum_{i=1}^{D} \sum_{j=1}^{D} \sum_{s=1}^{S} K_{i,j,s,t} X_{h_i,w_j,s}
\end{equation}
where $h_i = h - d / 2 + i,\; w_j = w - d / 2 + j$. 
The rank-($R_1$, $R_2$, $R_3$, $R_4$) Tucker decomposition \cite{de2000multilinear} has the form:
\begin{equation}
K_{i,j,s,t} = \sum_{r_1=1}^{R_1} \sum_{r_2=1}^{R_2} \sum_{r_3=1}^{R_3} \sum_{r_4=1}^{R_4} C_{r_1,r_2,r_3,r_4} U^{1}_{i,r_1} U^{2}_{j,r_2} U^{3}_{s,r_3} U^{4}_{t,r_4},
\end{equation}
where factor matrices $U^{n}$ are derived from mode-n unfolded kernel tensor, $C$ is the core tensor, which can be seen as a compressed version of original tensor.  

Most convolution kernels have special dimensions, and compressing a kernel tensor along special dimension does not give much difference.  Therefore, our method utilizes Tucker-2 decomposition  \cite{kim2015compression} for compression.  Tucker-2 decomposition is a variant of TKD, assuming partial factor matrices in TKD are identity matrices.  It reduces the computational cost for decomposition and usually allows a faster convergence to local minimum \cite{kolda2009tensor}.  In the case of decomposing a kernel tensor of order four, both factor matrices of special dimension are set to be identity matrices.  The rank-($D$, $D$, $R_3$, $R_4$) Tucker-2 decomposition has the form:
\begin{equation}
K_{i,j,s,t} = \sum_{r_3=1}^{R_3} \sum_{r_4=1}^{R_4} C_{i,j,r_3,r_4} U^{3}_{s,r_3} U^{4}_{t,r_4}
\end{equation}
where $U^{3}$ and $U^{4}$ are the factor matrices, $C \in R^{D\times D\times R_{3}\times R_{4}}$ is the core tensor.

TKD can be seen as a feature extractor for the kernel tensor $K \in R^D{}^\times{}^D{}^\times{}^S{}^\times{}^T{}$. The factored matrix $U^{3} \in R^S{}^\times{}^{R_3}$ represents a linear operator, which transforms the input feature maps into a compact one, so that the size of feature maps can reduced.  After the transformation, a compressed feature extractor $C \in R^{D\times D\times R_{3}\times R_{4}}$ is able to extract from the slimmer input feature maps, which therefore achieves acceleration.  Similarly,  the factored matrix $U^4 \in R^{{R_4} \times T}$ is used to transform the vectors in $R^{R_4}$ back to $R^T$, so it can connect the next layer. 

\subsection{Training}
\label{sec:training}
After the decomposition, our method keeps the decomposed model in full rank and continues training it.  Since the number of parameters in the decomposed model is more than the original one, such training usually gives a higher accuracy than the original one.  This step can make the later pruning more stable, because the pruning decision is based on the parameters and the structure of the decomposed form, not the original model.  

\begin{figure}
\centering
\includegraphics[scale = 0.19, clip]{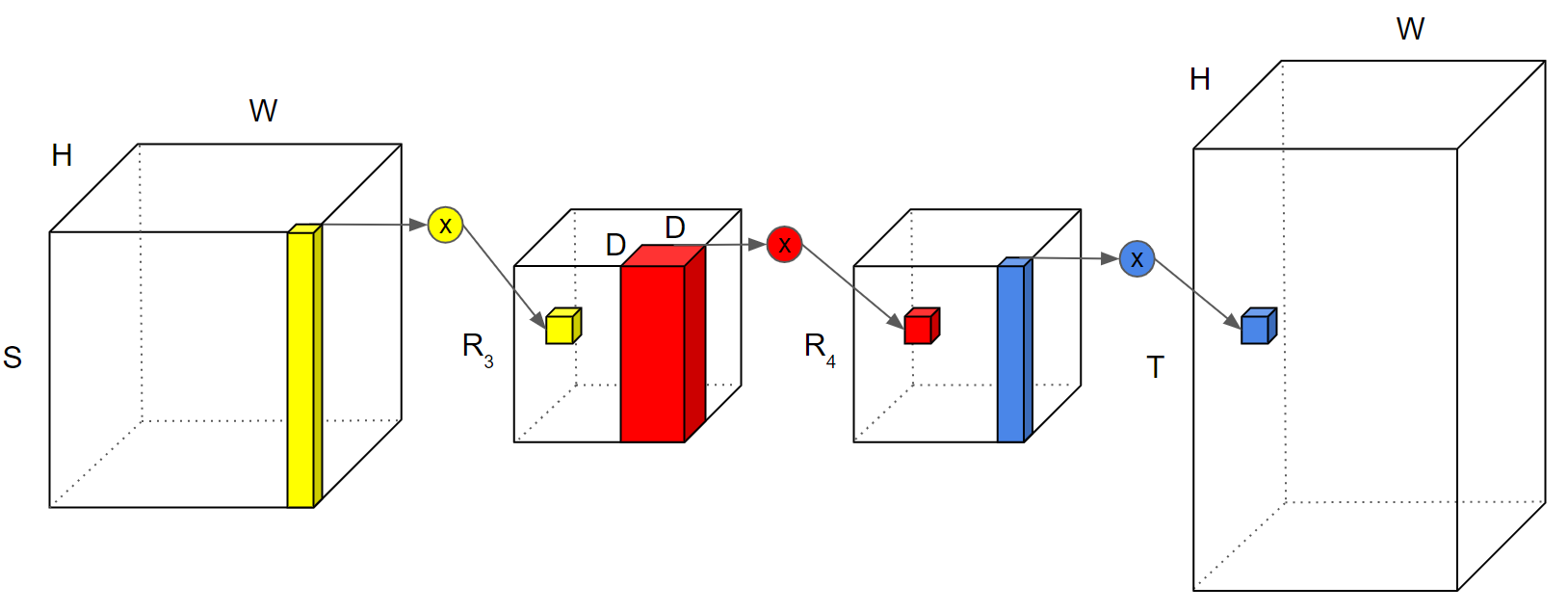}
\caption{Tucker decomposition on a kernel tensor with gates, where $R_3$ is the mode-3 rank and $R_4$ is the mode-4 rank.  The circles with a $\otimes$ sign means gates. All gates are colored the same as their kernels. $S$ is the number of input feature maps; $T$ is the number of output feature maps; $D$ is the special dimension of convolution kernel tensor; $H$ is the height of input feature map; $W$ is the width of input feature map.}
\label{fig:tkd}
\end{figure}

Before the training process, our method further factorizes each factor $v$ into
\begin{equation}
\label{eq:gate}
    v = \gamma_v \tilde{v}
\end{equation}
where $\gamma_v$ is a scalar, and $\tilde{v}$ is normalized to be a unit vector.  Geometrically, the scalar $\gamma_v$ is the magnitude of $v$, and $\tilde{v}$ is the direction of $v$.  The parameter $\gamma_v$ is called the \emph{gate} of $v$ since it will be used to represent the importance of the factor $v$ later. More specifically, let $G^{3} \in R^{R_3}$, $G^{4} \in R^T{}$ and $G^{C} \in R^{R_4}$ be the gates of the kernels in Tucker-2 decomposed kernel tensor. Their initial values are given as:
\begin{equation}
G^{C}_{r_4} = \|C_{:,:,:,r_4}\|,\; G^{3}_{r_3} = \|U^{3}_{:,r_3}\|,\; G^{4}_{t} = \|U^{4}_{t,:}\|
\end{equation}
With gates, the evaluation of $Y$ can be expressed as following equation,
\begin{align*}
Y_{h,w,t} &= G^{4}_{t} \sum_{r_4=1}^{R_4} U^{4}_{t,r_4} Z'_{h,w,r_4}\\
Z'_{h,w,r_4} &= G^{C}_{r_4} \sum_{i=1}^{D} \sum_{j=1}^{D} \sum_{r_3=1}^{R_3} C_{i,j,r_3,r_4} Z_{h_i,w_j,r_3}\\
Z_{h,w,r_3} &= G^{3}_{r_3} \sum_{s=1}^{S} U^{3}_{s,r_3} X_{h,w,s}
\end{align*}
The structure is also illustrated in Figure \ref{fig:tkd} pictorially. The training process includes all the gates as parameters, and keeps the multiplied factors normalized.  The loss function  used in here is just the classification loss.

\subsection{Compression}
After the training, the gates introduce in (\ref{eq:gate}) are combined with the regularization technique to determine the importance in of factors. Regularization is frequently used to avoid over-fitting. It is also commonly used in the channel pruning methods to measure the importance of network structures. Here, we use it to reveal the importance of factors, so our method can know exactly how many and which set of factors should be removed. 

The gate value is computed as the two-norm of its factor.  During the training process all factors inside decomposed kernels are restricted to be normalized. 
In the decomposition stage, our method adds a regularization term for those gates in the loss function, 
\begin{equation}
\label{eq:compress}
L = L_{\mathrm{class}} + \lambda L_{\mathrm{reg}}
\end{equation}
where $L_{\mathrm{class}}$ is classification loss, $L_{\mathrm{reg}}$ is regularization penalty, and $\lambda$ is a hyper-parameter that controls the ratio between two losses.

\begin{figure}[tb]
\centering
\includegraphics[scale = 0.25, clip]{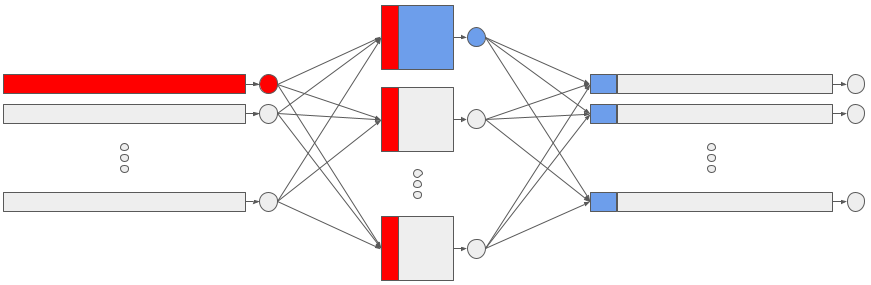}
\caption{Illustration of the process removing redundant kernels in the decomposed convolution layer. The first and third convolution layers have factors of special dimension one. The circles in front of factors are their gates respectively. Removing factors in the first convolution layer results in the removal of all weights in red; removing factors in the third convolution layer results in the removal of all weights in blue.}
\label{fig:remove_weight}
\end{figure}

After the training with the regularization on gates, our method removes the kernels whose gate value is close to zero, since their influence to the final accuracy is small.  The process of removing weights is illustrated in Figure \ref{fig:remove_weight}.  

\subsubsection{Funnel function}
\label{sec:funnel}
Two commonly used regularization methods are L2 and L1 functions. However, empirically we found that they cannot effectively produce the gate values during the compression. The L2 function, $F_2(x) = x^2$, pushes most of the value toward zero, but it fails to move most values to exact 0 at the end of training.  Removing those values in the pruning process can introduce a great drop in accuracy, which is hard to be recovered from later fine-tuning.  The L1 function, $F_1(x) = | x |$, also pushes $x$ toward zero, but the derivatives of all elements are either 1 or -1, except the zero elements.  As the result, most gates move to 0 at the same speed, including the important ones.  


To make the pruning decision easier, 
We define a new type of regularization. First, we define the funnel functions  as 
\begin{equation}
\label{eq:funnel}
F(x) = \frac{|x|}{c + |x|},
\end{equation}
where $x$ is the value of parameter in network, and $c$ is a positive hyper-parameter in $\mathbb{R}$. The idea is to push most value of parameters to exact zero rather than near it.  Therefore removing parameters of value zero does not cause a great drop in accuracy.  

The funnel function has two advantages over L1 and L2 regularization.  First, it can be more effective to distinguish the important and undesired factors.  Unlike L1 regularization whose derivative are either 1 or -1, the derivative of  $F$ is
\begin{equation}
\label{eq:funnel diff}
F'(x) = \begin{cases}
c / (c + x)^2 & \mbox{if } x\ge 0 \\
-c / (c - x)^2 & \mbox{if } x< 0.
\end{cases}
\end{equation}
As can be seen, for a constant $c$, if $x$ is closer to 0, $|F'(x)|$ is approaching to 1.  On the other hand, for a large $|x|$, $F'(x)$ is small. So that if $x$ falls near to zero, it is push to zero; and if $x$ is large enough, its value is not changed much during the regularization.   As the result, the values closer to zero would become exact zero at the end of compression. 

\begin{figure}[tb]
\centering
\includegraphics[scale = 0.5, clip]{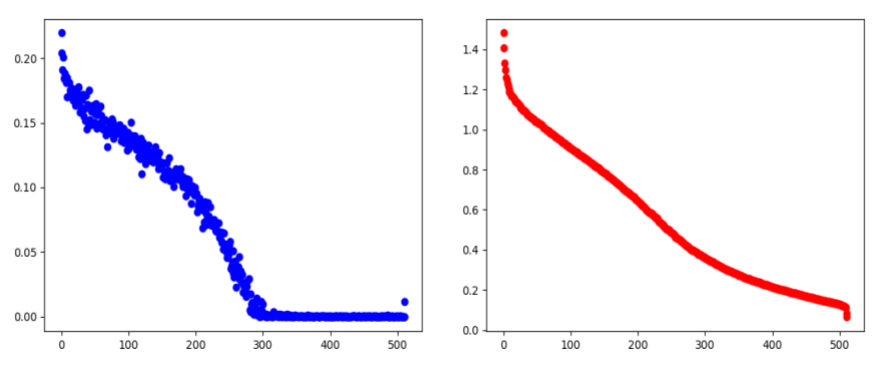}
\caption{Illustration of the value of gates before and after training process.  Values of gates before training are on the right, and after training are on the left. Gate values are sorted in the descending order along x-axis. Y-axis means the values of gates.}
\label{fig:before_after_1}
\end{figure}

The second advantage of using funnel function is the compressed network does not require heavier re-training after pruning.  To the end of the training, the norms of unimportant factors, equal to the gate values, are almost zeros, which means those factors mostly have no influence on the network accuracy.  The effect of applying funnel regularization is shown in Figure \ref{fig:before_after_1}. As can be seen, the gates of value close to zero are pushed to zero after training, but the value of important parameters are reserved.  However, the figure also shows that the funnel function still cannot create a clear gap between the important and the redundant gate values.  So our method still needs a pruning threshold to decide the cut point.  However, comparing to other methods, the range of the uncertain values is much smaller, as can be seen from the figure.

The parameter $c$ in the funnel function plays an important role. Our study finds that a varied $c$ could give a better accuracy than a constant assignment, which will be shown in the experiments.  The idea is to push all values of gates toward zero at the first few epochs, by which the funnel function looks like the L1 function. Next, the value of $c$ is gradually reduced to accelerate the smaller gates becoming to zero.  Meanwhile, gates with large values, which often are important parameters, can keep their values.

Here we proposed two ways to adjust the parameter $c$. The first is by linear decay,
\begin{equation}
\label{eq:schedule}
c = c_1 + (c_2-c_1) \times i/n
\end{equation}
where $c_1$ is the initial value of $c$, $c_2$ is the final value of $c$, $n$ is the number of total epochs.  The second one is exponential decay, which multiplies  a factor of $\sigma$ in every $m$ epochs. 

\subsection{Fine-tuning}
Not all decomposed kernel tensors become smaller after pruning.  After compression, the computational cost of decomposed kernel tensor can be calculated.  For the original tensor, its computational cost is
\begin{equation}
C_{\mathrm{original}} = HWD^2ST
\end{equation}
where $S$ is the number of input channel; $T$ is the output channel; $H$ and $W$ are the height and width of input feature map respectively; $D$ is the size of convolution kernel.
For the decomposed kernel tensor, it is 
\begin{equation}
C_{\mathrm{decomposed}} = HW(SR_3 + R_3D^{2}R_4 + R_4T)\;
\end{equation}
where $R_3$ and $R_4$ are the ranks of the first and the third factors.

For a factor of a larger gate value, our method compares the computational cost of a decomposed kernels and the computational cost of its original tensor.  If the computational cost of the decomposed kernel is less than that of the original one, our method multiplies the gate value back to its kernel and removes gates from the decomposed factors.
Otherwise, we reverse the gating and decomposition to construct the original tensor.  With this comparison added after compression process, our proposed method can automatically decide which kernel tensor is to be decomposed or not. Moreover,  all the gates added into model only exist in compression process, which does not introduce no extra parameter to the final model. 

After that our method retrains the adjusted model since the model structures are different. The used loss function simply is the classification loss.


\section{Experiments}
\label{chap:exp}

Our experiments use Imagenet2012 dataset, and run on a single NVIDIA Tesla V100 GPU card.  The used platform is pytorch version 20.08 and python version 3.3.  We evaluated our method on ResNet18, ResNet50 and DenseNet121.  Without further specification, the used tensor decomposition is TKD; the learning rate used for decomposition is $10^{-3}$; the parameter $c$ used in the funnel function is exponentially decay from $1$ to $10^{-4}$.  The pruning threshold for ResNet18 is $10^{-3}$, and for ResNet50 and DenseNet121 is $10^{-4}$. 

The measurement of accuracy is Top-1. Another performance measurement is the theoretical speed up, which is defined as 
\begin{equation}
\mbox{Speed up}=\frac{\mbox{GMAC of compressed model}}{\mbox{GMAC of pre-trained model}},
\label{eqn:speedup}
\end{equation}
where GMAC is the gigaMACs of a model, forwarding a single 224x224x3 image as the input.  The term $\#$Param means the number of parameters in million.  

We have three sets of experiments.  The first one compares our method with others.  The second one performs ablation tests.  The third one evaluates the effectiveness of funnel function. 

\subsection{Comparison with other methods}
We compared our method with other methods for model compression on Resnet18.  We measured their Top-1 accuracy and the GMAC after pruning, which are listed in Table \ref{tab:comparison}.  Most results of Top-1 and GMAC are retrieved from their papers, except MUSCO \cite{gusak2019automated} and CP-TPM \cite{astrid2017cp}, whose results are  obtained from our execution, since they are relatively new and their codes are publicly available.  The reduction rate in MUSCO is set to $0.5$ in this experiment. For the method introduced in \cite{stable2020}, although the results shown in the paper are great, we cannot reproduce them since its source code is not in open to public. 

We have also conducted experiment for GAS-of-EVBMF \cite{kim2015compression}, but the rank selected by GAS-of-EVBMF is too aggressive, leading to a extremely small compressed model that is hard to compared to.  We will present the result of TKD-VBMF separately.  

\begin{table}
\small
\caption{Comparison with other methods.}
\centering
    \begin{tabular}{ | c | c | c | c | c | } 
    \hline
    Method & Top-1 & \centering{$\Delta$Top-1} & GMAC & speed up \\ 
    \hline
      baseline  & 69.76 & 0.00 & 1.82 & 1.00 \\ 
    \hline
     SlimNet  & 67.99 & -1.76 & 1.31 & 1.39 \\ 
    \hline
     LCL  & 66.11 & -3.65 & 1.19 & 1.53 \\ 
    \hline
     CGNN  & 68.14 & -1.62 & 1.13 & 1.61 \\ 
    \hline
     FLP  & 66.58 & -3.18 & 1.06 & 1.72 \\ 
    \hline
     DCP  & 67.47 & -2.29 & 0.96 & 1.89 \\ 
    \hline
     FBS  & 67.22 & -2.54 & 0.92 & 1.98 \\ 
    \hline
    CP-TPM  & 67.51 & -2.25 & 1.15 & 1.58 \\ 
    \hline
    MUSCO  & 68.94 & -0.82 & 1.07 & 1.70 \\ 
    \hline
    Stable Low-rank  & 69.07 & -0.69 & 0.59 & 3.09 \\ 
    \hline
     Ours & 69.04 & -0.72 & 0.90 & 2.02 \\
    \hline
    \end{tabular}
\label{tab:comparison}
\end{table}


\subsubsection{Comparison with TKD-VBMF}

TKD-VBMF\cite{kim2015compression} accelerates  CNNs using TKD with approximate ranks determined by  GAS-of-EVBMF on mode-n unfolded kernel tensor \cite{kolda2009tensor}. Note that GAS-of-EVBMF is also applied on mode-n unfolded kernel tensor to define the extreme rank in MUSCO \cite{gusak2019automated}.  We compared our method with TKD-VBMF.  The compression process approximates kernel tensors with the ranks given by VBMF, so the lower the rank, the higher portion of kernel tensors is removed.  Therefore one can get a smaller and faster model. But the compressed model often suffers from a huge drop in accuracy.  

\begin{table}
\small
\caption{Comparison with TKD-VBMF.}
\centering
    \begin{tabular}{ | c | c | c | c | c | } 
    \hline
    Method & Top-1 & $\Delta$Top-1 & GMAC & speed up  \\ 
    \hline
     TKD-VBMF 
     & 61.06 & -8.70 & 0.38 & 4.79 \\ 
    \hline
    Ours & 61.67 & -8.09 & 0.36 & 5.06 \\ 
    \hline
    \end{tabular}
\label{tab:tkd-vbmf}
\end{table}

Table \ref{tab:tkd-vbmf} lists the experimental results on ResNet18.  To match the model size obtained by TKD-VBMF, we also pruned more parameters than the original settings.  As can be seen, the compressed model obtained by our method can still get better accuracy than that of TKD-VBMF.   

\subsubsection{Comparison with CP-TPM}
We compared our method with CP-TPM \cite{astrid2017cp} on on ResNet18, ResNet50, and DenseNet121, since CP-TPM is the only method with publicly available code that can compress large models effectively. Instead of approximating a kernel tensor at once, CP-TPM iteratively approximates the kernel tensor in a greedy approach using CP decomposition (CPD).  In CP-TPM, the approximation rank of each kernel tensor is determined with a loss distribution.  Only an approximate rank is required for a model.  The approximate rank of each kernel tensor is then determined by the portion of the loss. The loss distribution is used to determine the approximation rank of each block in CP-TPM, where the higher portion of loss a kernel tensor contributes to total loss, the larger rank is given to the kernel. Rank here is used to determine the loss distribution, not the approximate rank used to decompose kernel tensors.

\begin{table}
\small
\caption{Comparison with CP-TPM on ResNet18, ResNet50, and DenseNet121.}
\centering
    \begin{tabular}{|c|c| c | c | c | } 
    \hline
   Model & Method & GMAC & $\#$Param & Top-1 \\ 
    \hline
    \multirow{2}{*}{ResNet18} & 
CP-TPM & 1.15 & 7.38 & 67.514 \\ 
    \cline{2-5}
    & Ours & 0.90 & 5.03 & 69.040 \\ 
    \hline
    \multirow{2}{*}{ResNet50} 
    &CP-TPM & 2.57 & 15.47 & 70.856 \\ 
    \cline{2-5}
    &Ours & 2.18 & 15.13 & 75.070 \\ 
    \hline
    \multirow{2}{*}{DenseNet50} 
    &CP-TPM & 1.88 & 6.09 & 71.541 \\ 
    \cline{2-5}
    &Ours & 1.56 & 5.57 & 74.144 \\ 
    \hline
    \end{tabular}
\label{tab:CPTPM}
\end{table}

The experimental results are shown in Table \ref{tab:CPTPM}.  As can be seen, our algorithm outperforms TP-TPM in terms of model accuracy and model size for all models.  In addition, for larger models, our method shows more advantage to compress the model without scarifying the accuracy.

\subsection{Ablation test}
We have studied the effect of using different decomposition methods, learning rate, and pruning threshold for our method.  The results are presented below.  The results of the ablation study on the funnel function are presented in the next section.

\subsubsection{Different decomposition methods}
\begin{table}
\small
\caption{Comparison of different decomposition methods on ResNet18.}
\centering
    \begin{tabular}{ | c | c | c | c |} 
    \hline
     Decomposition & Top-1 & GMAC & $\#$Param \\ 
    \hline
    SVD & 67.122 & 1.25 & 7.53 \\
    \hline
    CPD & 66.214 & 1.08 & 5.32 \\
    \hline
    TKD & 67.368 & 0.82 & 4.62 \\ 
    \hline
    \end{tabular}
\label{tab:different_decomposition}
\end{table}

Our method can be applied to various low-rank approximation methods.  Here we evaluated the performance of different decomposition methods, including SVD, TKD and CPD, integrated into our framework. We measured these decomposition methods by their Top-1 accuracy, GMAC, and $\#$Param of the compressed models for ResNet18. For easier comparison, we adjusted the values of GMAC and $\#$Param so their Top-1 accuracy are similar. 
In this experiment all the compressed models are tuned for 50 epochs after compression. The learning rate in this experiment is set to $10^{-2}$.  

The results of comparison are listed in the Table \ref{tab:different_decomposition}.  As can be seen, the model processed with TKD has the best Top-1 accuracy and the smallest size. This result is consistent with the common belief that CPD is better for feature extraction, and TKD is suitable for model compression \cite{kolda2009tensor}. For SVD, although the model accuracy is also well maintained, the model size is hard to be reduced owing the high matrix rank.

\subsubsection{Learning rate}
Keeping learning rate small is very important in the decompression stage, since there is no non-linear activation function between decomposed kernels, so a decomposed model is prone to the issue of gradient explosion and gradient vanishing, especially for CPD \cite{de2008tensor} \cite{astrid2017cp}. 

We compared different learning rates, $10^{-2}$, $10^{-3}$, and $10^{-4}$, in the compression stage. The model for evaluation is ResNet18.  We trained models with different learning rates and measured their Top-1 accuracy in the first epoch and the highest Top-1 accuracy in 50 epochs. 

The results are  summarized in Table \ref{tab:different_lr}, which shows that a larger learning rate could cause a great accuracy drop. In our experiment, most model experience a accuracy drop in the beginning of the compression process. The larger the learning rate the lower Top-1 accuracy in the beginning we get. On the other hand, the learning rate cannot be too small either, because the loss would stagnate quickly. To find a proper learning rate during the training require further researches.  

\begin{table}
\small
\caption{Comparison of different learning rates in the compression stage on ResNet18.}
\centering
    \begin{tabular}{ | c | c | c |} 
    \hline
    L-rate& \centering{Top-1 (epoch 1)} & Top-1 (epoch 50) \\ 
    \hline
    $10^{-2}$ & 8.530 & 63.592 \\ 
    \hline
    $10^{-3}$ & 49.952 & 67.842 \\ 
    \hline
    $10^{-4}$ & 57.060 & 67.594 \\ 
    \hline
    \end{tabular}
\label{tab:different_lr}
\end{table}

\subsubsection{Pruning threshold}
In the compression stage, our method needs a magnitude to decide how much factors and which factors should be removed.  The conventional belief is that the more parameters are pruned, the less accuracy can be acquired.  However, the real situation can be more complex, because it also depends on the network structures and training data.  

\begin{table}
\small
\caption{Compressed results for different pruning threshold.}
\centering
    \begin{tabular}{|c | c | c | c | c | } 
    \hline
    Model & Threshold & GMAC & $\#$Param & Top-1 \\ 
    \hline
    \multirow{3}{*}{ResNet18} &
    baseline & 1.82 & 11.69 & 69.76 \\ 
    \cline{2-5}
    & 0.01 & 0.96 & 5.32 & 68.98 \\ 
    \cline{2-5}
    & 0.02 & 0.90 & 5.03 & 69.04 \\ 
    \hline
    \multirow{3}{*}{ResNet50} &
    baseline & 4.12 & 25.56 & 76.15 \\ 
    \cline{2-5}
    & 0.01 & 2.18 & 15.13 & 75.07 \\ 
    \cline{2-5}
    & 0.02 & 1.64 & 11.74 & 74.31 \\ 
    \hline
    \multirow{3}{*}{DenseNet121} &
    baseline & 2.88 & 7.98 & 74.65 \\ 
    \cline{2-5}
    & 0.01 & 1.86 & 6.27 & 75.26 \\ 
    \cline{2-5}
    & 0.02 & 1.56 & 5.57 & 74.14 \\ 
    \hline
    \end{tabular}
\label{tab:different_Resnet18}
\end{table}

The experimental results on ResNet18 are listed in Table \ref{tab:different_Resnet18}. For each model, there are three configurations to compare.  The first one is the baseline; the second one is a smaller threshold; and the third one is a bigger one.   For ResNet18, the network with the smaller threshold, whose model size is larger, does not deliver a higher accuracy. On the other hand, the larger threshold, which prunes more parameters, has a higher accuracy. This is a counter-intuition phenomenon, but their values are close to each other.  Similar case also happens for DenseNet121 baseline and DenseNet121 with pruning threshold $10^{-3}$, in which the smaller model can achieve a higher accuracy.

\subsection{Funnel function}
\subsubsection{Different regularization functions}
This experiment compares the effectiveness of the funnel function with L1 and L2 regularization.  The model to prune is ResNet18, which is trained for 50 epochs and pruned with the same threshold value $10^{-3}$ after training.  Any factor in the decomposed kernel tensor with its two-norm less than the threshold value is considered to be redundant and will be removed.  After pruning there is no fine-tuning performed.

\begin{table}
\small
\caption{Comparison of different regularization functions.}
\centering
    \begin{tabular}{ | c | c | c | c |} 
    \hline
    & Top-1 & GMAC & $\#$Param \\ 
    \hline
    baseline & 69.732 & 1.80 & 11.55 \\
    \hline
    L1 regularization & 67.458 & 0.98 & 6.31 \\ 
    \hline
    L2 regularization & 67.914 & 1.78 & 11.44 \\ 
    \hline
    funnel regularization & 68.242 & 0.83 & 5.53 \\ 
    \hline
    \end{tabular}
\label{tab:different_regularization}
\end{table}

Table \ref{tab:different_regularization} compares the effectiveness of different regularization methods.  It lists Top-1, GMAC and $\#$Param of each model after pruning.  The first line is the baseline ResNet18 model trained without regularization.  The results show that the proposed funnel function outperforms the other two commonly used regularization functions.  For L2 function, it cannot effectively reduce the number of parameters.  For L1 function, it suffers the accuracy drop.

\subsubsection{Parameter $c$}
The parameter $c$ in funnel function, $F(x) = |x| / (|x| + c)$, plays a critical role.  In Section
\ref{sec:funnel}, we proposed three different ways to setup the parameter $c$. One is the constant assignment; one is linear decay; and one is exponential decay.  This experiment evaluates the effectiveness of different ways to setup $c$. 


The experimental results are listed in Table \ref{tab:schedule_c}. For the constant assignment,  the value of $c$ is ranged from $10^{2}$ to $10^{-3}$. For the linear decay (adjust), $c$ is set to decay from $10$ to $10^{-3}$ in 100 epochs.  For the exponential decay, the initial $c$ is $10^2$, and multiplied by $0.1$ in every 5 epochs.  We measure their effectiveness by Top-1 accuracy and the pruning ratio, which is the percentage of the kernels whose gate values are smaller than the pruning threshold.  For constant assignment, the results show a trade-off between pruning ratio and the Top-1 accuracy.  When $c$ is large $10^2$ or small $10^{-3}$, the pruning ratio is small, but the Top-1 accuracy is high.  When $c$ equals to $1$ or $0.1$, the pruning ratio is high, but the Top-1 accuracy drop significantly.

\begin{table}
\small
\caption{Comparison of different assignment of $c$.}
\centering
    \begin{tabular}{ | c | c | c |} 
    \hline
    C values & Pruning ratio & Top-1 \\ 
    \hline
    $c = 10^{2}$ & 0.3\% & 70.422 \\
    \hline
    $c = 10^{1}$ & 25.0\% & 69.224 \\
    \hline
    $c = 10^{0}$ & 71.1\% & 64.038 \\
    \hline
    $c = 10^{-1}$ & 84.2\% & 52.486 \\
    \hline
    $c = 10^{-2}$ & 13.8\% & 69.780 \\
    \hline
    $c = 10^{-3}$ & 1.3\% & 70.426 \\
    \hline
    linear decay & 41.8\% & 69.204 \\
    \hline
    exp. decay & 21.7\% & 66.574 \\
    \hline
    $c = | g_{0} |$ & 53.2\% & 69.316 \\
    \hline
    \end{tabular}
\label{tab:schedule_c}
\end{table}

This phenomenon can be reasoned as follows. For $|x|\le 1$, when $c$ is large, $F \sim |x|/c$, whose effect is similar to the L1 regularization. When $c$ is small, $F \sim |x|/|x| = 1$ for $|x| >> c$.  Thus, it has no any influence on $x$ except the really small ones.  For median $c$, the value of $x$ whose initial value in the range $[-1, 1]$ converges to 0 efficiently.  However, the side-effect is that the values of some important gates are also becoming to 0, which results a big accuracy drop. 

In general, the dynamic value of $c$ performs better than the constant assignment.  In the beginnings, the L1 regularization should be used to distinguish the importance of each factors, so a larger $c$ should be used.  Later, the regularization should accelerate the convergence of the small gate values.  Therefore, a smaller $c$ is better.  

The experiments also indicate that the result of the linear decay is better than that of the exponential decay in terms of pruning ratio and the Top-1 accuracy.  The possible reason is the exponential decay may reduce the parameter $c$ too aggressively so some important parameters are also pull down to zero in the early stage.

\section{Conclusion and Future Work}
In this paper, we presented a model compression method based on the low-rank approximation.  Unlike the previous methods, our method does not compress the network layer by layer, but utilizes the optimization method to determine the pruning strategy for all layers simultaneously.  Moreover, we proposed a new compression flow, which trains the decomposed model first before pruning.  Last, a new regularization function, called funnel function, is proposed to make the rank selection easier.  Experiments on various models show the effectiveness of the proposed method.  Comparing to other existing methods, our algorithm has the good speedup and the smallest accuracy drop.

There are many directions to explore in the future. First, we can combine our method with the stabilization approach \cite{stable2020}, since it still uses bruit force method for rank selection.  Second, the proposed method could be extended to other types of layers which can be formulated as tensors, or other types of network, such as transformer \cite{Attention} or GPT-3 \cite{GPT-3}.  Third, our experiments showed the power of using the funnel function.  However, its properties may require more studied, such as how to adjust the parameter $c$ adaptively. Last, the theoretical analysis on the trade-off between model size and model accuracy is an important research, since it would give us a solid ground to develop better model compression methods.

\bibliography{egbib}

\begin{thebibliography}{41}
\providecommand{\natexlab}[1]{#1}

\bibitem[{Astrid and Lee(2017)}]{astrid2017cp}
Astrid, M.; and Lee, S.-I. 2017.
\newblock Cp-decomposition with tensor power method for convolutional neural
  networks compression.
\newblock In \emph{2017 IEEE International Conference on Big Data and Smart
  Computing (BigComp)}, 115--118. IEEE.

\bibitem[{Brown et~al.(2020)Brown, Mann, Ryder, Subbiah, Kaplan, Dhariwal,
  Neelakantan, Shyam, Sastry, Askell, Agarwal, Herbert-Voss, Krueger, Henighan,
  Child, Ramesh, Ziegler, Wu, Winter, Hesse, Chen, Sigler, Litwin, Gray, Chess,
  Clark, Berner, McCandlish, Radford, Sutskever, and Amodei}]{GPT-3}
Brown, T.; Mann, B.; Ryder, N.; Subbiah, M.; Kaplan, J.~D.; Dhariwal, P.;
  Neelakantan, A.; Shyam, P.; Sastry, G.; Askell, A.; Agarwal, S.;
  Herbert-Voss, A.; Krueger, G.; Henighan, T.; Child, R.; Ramesh, A.; Ziegler,
  D.; Wu, J.; Winter, C.; Hesse, C.; Chen, M.; Sigler, E.; Litwin, M.; Gray,
  S.; Chess, B.; Clark, J.; Berner, C.; McCandlish, S.; Radford, A.; Sutskever,
  I.; and Amodei, D. 2020.
\newblock Language Models are Few-Shot Learners.
\newblock In Larochelle, H.; Ranzato, M.; Hadsell, R.; Balcan, M.~F.; and Lin,
  H., eds., \emph{Advances in Neural Information Processing Systems},
  volume~33, 1877--1901. Curran Associates, Inc.

\bibitem[{Carroll and Chang(1970)}]{carroll1970analysis}
Carroll, J.~D.; and Chang, J.-J. 1970.
\newblock Analysis of individual differences in multidimensional scaling via an
  N-way generalization of “Eckart-Young” decomposition.
\newblock \emph{Psychometrika}, 35(3): 283--319.

\bibitem[{Chen et~al.(2015)Chen, Wilson, Tyree, Weinberger, and
  Chen}]{chen2015compressing}
Chen, W.; Wilson, J.; Tyree, S.; Weinberger, K.; and Chen, Y. 2015.
\newblock Compressing neural networks with the hashing trick.
\newblock In \emph{International conference on machine learning}, 2285--2294.
  PMLR.

\bibitem[{Cheng et~al.(2015)Cheng, Felix, Feris, Kumar, Choudhary, and
  Chang}]{cheng2015fast}
Cheng, Y.; Felix, X.~Y.; Feris, R.~S.; Kumar, S.; Choudhary, A.; and Chang,
  S.-F. 2015.
\newblock Fast neural networks with circulant projections.
\newblock \emph{arXiv preprint arXiv:1502.03436}, 2.

\bibitem[{Cheng et~al.(2017)Cheng, Wang, Zhou, and Zhang}]{cheng2017survey}
Cheng, Y.; Wang, D.; Zhou, P.; and Zhang, T. 2017.
\newblock A survey of model compression and acceleration for deep neural
  networks.
\newblock \emph{arXiv preprint arXiv:1710.09282}.

\bibitem[{Choi, El-Khamy, and Lee(2016)}]{choi2016towards}
Choi, Y.; El-Khamy, M.; and Lee, J. 2016.
\newblock Towards the limit of network quantization.
\newblock \emph{arXiv preprint arXiv:1612.01543}.

\bibitem[{De~Lathauwer, De~Moor, and
  Vandewalle(2000{\natexlab{a}})}]{de2000multilinear}
De~Lathauwer, L.; De~Moor, B.; and Vandewalle, J. 2000{\natexlab{a}}.
\newblock A multilinear singular value decomposition.
\newblock \emph{SIAM journal on Matrix Analysis and Applications}, 21(4):
  1253--1278.

\bibitem[{De~Lathauwer, De~Moor, and
  Vandewalle(2000{\natexlab{b}})}]{de2000best}
De~Lathauwer, L.; De~Moor, B.; and Vandewalle, J. 2000{\natexlab{b}}.
\newblock On the best rank-1 and rank-(r 1, r 2,..., rn) approximation of
  higher-order tensors.
\newblock \emph{SIAM journal on Matrix Analysis and Applications}, 21(4):
  1324--1342.

\bibitem[{De~Lathauwer and Nion(2008)}]{blockTKD}
De~Lathauwer, L.; and Nion, D. 2008.
\newblock Decompositions of a Higher-Order Tensor in Block Terms—Part III:
  Alternating Least Squares Algorithms.
\newblock \emph{SIAM Journal on Matrix Analysis and Applications}, 30(3):
  1067--1083.

\bibitem[{De~Silva and Lim(2008)}]{de2008tensor}
De~Silva, V.; and Lim, L.-H. 2008.
\newblock Tensor rank and the ill-posedness of the best low-rank approximation
  problem.
\newblock \emph{SIAM Journal on Matrix Analysis and Applications}, 30(3):
  1084--1127.

\bibitem[{{Deng} et~al.(2020){Deng}, {Li}, {Han}, {Shi}, and
  {Xie}}]{Deng2020survey}
{Deng}, L.; {Li}, G.; {Han}, S.; {Shi}, L.; and {Xie}, Y. 2020.
\newblock Model Compression and Hardware Acceleration for Neural Networks: A
  Comprehensive Survey.
\newblock \emph{Proceedings of the IEEE}, 108(4): 485--532.

\bibitem[{Denil et~al.(2013)Denil, Shakibi, Dinh, Ranzato, and
  De~Freitas}]{denil2013predicting}
Denil, M.; Shakibi, B.; Dinh, L.; Ranzato, M.; and De~Freitas, N. 2013.
\newblock Predicting parameters in deep learning.
\newblock \emph{arXiv preprint arXiv:1306.0543}.

\bibitem[{Denton et~al.(2014)Denton, Zaremba, Bruna, LeCun, and
  Fergus}]{denton2014exploiting}
Denton, E.; Zaremba, W.; Bruna, J.; LeCun, Y.; and Fergus, R. 2014.
\newblock Exploiting linear structure within convolutional networks for
  efficient evaluation.
\newblock \emph{arXiv preprint arXiv:1404.0736}.

\bibitem[{Dong et~al.(2017)Dong, Huang, Yang, and Yan}]{dong2017more}
Dong, X.; Huang, J.; Yang, Y.; and Yan, S. 2017.
\newblock More is less: A more complicated network with less inference
  complexity.
\newblock In \emph{Proceedings of the IEEE Conference on Computer Vision and
  Pattern Recognition}, 5840--5848.

\bibitem[{Eld{\'e}n and Savas(2009)}]{elden2009newton}
Eld{\'e}n, L.; and Savas, B. 2009.
\newblock A Newton--Grassmann Method for Computing the Best Multilinear
  Rank-(r\_1, r\_2, r\_3) Approximation of a Tensor.
\newblock \emph{SIAM Journal on Matrix Analysis and applications}, 31(2):
  248--271.

\bibitem[{Gao et~al.(2018)Gao, Zhao, Dudziak, Mullins, and Xu}]{gao2018dynamic}
Gao, X.; Zhao, Y.; Dudziak, {\L}.; Mullins, R.; and Xu, C.-z. 2018.
\newblock Dynamic channel pruning: Feature boosting and suppression.
\newblock \emph{arXiv preprint arXiv:1810.05331}.

\bibitem[{Gong et~al.(2014)Gong, Liu, Yang, and Bourdev}]{gong2014compressing}
Gong, Y.; Liu, L.; Yang, M.; and Bourdev, L. 2014.
\newblock Compressing deep convolutional networks using vector quantization.
\newblock \emph{arXiv preprint arXiv:1412.6115}.

\bibitem[{Gusak et~al.(2019)Gusak, Kholiavchenko, Ponomarev, Markeeva,
  Blagoveschensky, Cichocki, and Oseledets}]{gusak2019automated}
Gusak, J.; Kholiavchenko, M.; Ponomarev, E.; Markeeva, L.; Blagoveschensky, P.;
  Cichocki, A.; and Oseledets, I. 2019.
\newblock Automated multi-stage compression of neural networks.
\newblock In \emph{Proceedings of the IEEE International Conference on Computer
  Vision Workshops}, 0--0.

\bibitem[{Han, Mao, and Dally(2015)}]{han2015deep}
Han, S.; Mao, H.; and Dally, W.~J. 2015.
\newblock Deep compression: Compressing deep neural networks with pruning,
  trained quantization and huffman coding.
\newblock \emph{arXiv preprint arXiv:1510.00149}.

\bibitem[{Harshman and Lundy(1994)}]{harshman1994parafac}
Harshman, R.~A.; and Lundy, M.~E. 1994.
\newblock PARAFAC: Parallel factor analysis.
\newblock \emph{Computational Statistics \& Data Analysis}, 18(1): 39--72.

\bibitem[{He, Zhang, and Sun(2017)}]{he2017channel}
He, Y.; Zhang, X.; and Sun, J. 2017.
\newblock Channel pruning for accelerating very deep neural networks.
\newblock In \emph{Proceedings of the IEEE International Conference on Computer
  Vision}, 1389--1397.

\bibitem[{Hua et~al.(2019)Hua, Zhou, De~Sa, Zhang, and Suh}]{hua2019channel}
Hua, W.; Zhou, Y.; De~Sa, C.~M.; Zhang, Z.; and Suh, G.~E. 2019.
\newblock Channel gating neural networks.
\newblock In \emph{Advances in Neural Information Processing Systems},
  1886--1896.

\bibitem[{Kapteyn, Neudecker, and Wansbeek(1986)}]{kapteyn1986approach}
Kapteyn, A.; Neudecker, H.; and Wansbeek, T. 1986.
\newblock An approach ton-mode components analysis.
\newblock \emph{Psychometrika}, 51(2): 269--275.

\bibitem[{Kim and Choi(2007)}]{kim2007nonnegative}
Kim, Y.-D.; and Choi, S. 2007.
\newblock Nonnegative tucker decomposition.
\newblock In \emph{2007 IEEE Conference on Computer Vision and Pattern
  Recognition}, 1--8. IEEE.

\bibitem[{Kim et~al.(2015)Kim, Park, Yoo, Choi, Yang, and
  Shin}]{kim2015compression}
Kim, Y.-D.; Park, E.; Yoo, S.; Choi, T.; Yang, L.; and Shin, D. 2015.
\newblock Compression of deep convolutional neural networks for fast and low
  power mobile applications.
\newblock \emph{arXiv preprint arXiv:1511.06530}.

\bibitem[{Kolda and Bader(2009)}]{kolda2009tensor}
Kolda, T.~G.; and Bader, B.~W. 2009.
\newblock Tensor decompositions and applications.
\newblock \emph{SIAM review}, 51(3): 455--500.

\bibitem[{Kroonenberg and De~Leeuw(1980)}]{kroonenberg1980principal}
Kroonenberg, P.~M.; and De~Leeuw, J. 1980.
\newblock Principal component analysis of three-mode data by means of
  alternating least squares algorithms.
\newblock \emph{Psychometrika}, 45(1): 69--97.

\bibitem[{Lebedev et~al.(2014)Lebedev, Ganin, Rakhuba, Oseledets, and
  Lempitsky}]{lebedev2014speeding}
Lebedev, V.; Ganin, Y.; Rakhuba, M.; Oseledets, I.; and Lempitsky, V. 2014.
\newblock Speeding-up convolutional neural networks using fine-tuned
  cp-decomposition.
\newblock \emph{arXiv preprint arXiv:1412.6553}.

\bibitem[{Liu et~al.(2017)Liu, Li, Shen, Huang, Yan, and
  Zhang}]{liu2017learning}
Liu, Z.; Li, J.; Shen, Z.; Huang, G.; Yan, S.; and Zhang, C. 2017.
\newblock Learning efficient convolutional networks through network slimming.
\newblock In \emph{Proceedings of the IEEE International Conference on Computer
  Vision}, 2736--2744.

\bibitem[{Luo, Wu, and Lin(2017)}]{luo2017thinet}
Luo, J.-H.; Wu, J.; and Lin, W. 2017.
\newblock Thinet: A filter level pruning method for deep neural network
  compression.
\newblock In \emph{Proceedings of the IEEE international conference on computer
  vision}, 5058--5066.

\bibitem[{Nakajima et~al.(2012)Nakajima, Tomioka, Sugiyama, and
  Babacan}]{nakajima2012perfect}
Nakajima, S.; Tomioka, R.; Sugiyama, M.; and Babacan, S. 2012.
\newblock Perfect dimensionality recovery by variational Bayesian PCA.
\newblock \emph{Advances in Neural Information Processing Systems}, 25:
  971--979.

\bibitem[{Novikov et~al.(2015)Novikov, Podoprikhin, Osokin, and
  Vetrov}]{novikov2015tensorizing}
Novikov, A.; Podoprikhin, D.; Osokin, A.; and Vetrov, D. 2015.
\newblock Tensorizing neural networks.
\newblock \emph{arXiv preprint arXiv:1509.06569}.

\bibitem[{Oseledets and Tyrtyshnikov(2009)}]{TT2009}
Oseledets, I.~V.; and Tyrtyshnikov, E.~E. 2009.
\newblock Breaking the Curse of Dimensionality, Or How to Use SVD in Many
  Dimensions.
\newblock \emph{SIAM Journal on Scientific Computing}, 31(5): 3744--3759.

\bibitem[{Phan et~al.(2020)Phan, Sobolev, Sozykin, Ermilov, Gusak,
  Tichavsk{\'y}, Glukhov, Oseledets, and Cichocki}]{stable2020}
Phan, A.-H.; Sobolev, K.; Sozykin, K.; Ermilov, D.; Gusak, J.; Tichavsk{\'y},
  P.; Glukhov, V.; Oseledets, I.; and Cichocki, A. 2020.
\newblock Stable Low-Rank Tensor Decomposition for Compression of Convolutional
  Neural Network.
\newblock In Vedaldi, A.; Bischof, H.; Brox, T.; and Frahm, J.-M., eds.,
  \emph{Computer Vision -- ECCV 2020}, 522--539. Cham: Springer International
  Publishing.

\bibitem[{Shashua and Hazan(2005)}]{shashua2005non}
Shashua, A.; and Hazan, T. 2005.
\newblock Non-negative tensor factorization with applications to statistics and
  computer vision.
\newblock In \emph{Proceedings of the 22nd international conference on Machine
  learning}, 792--799.

\bibitem[{Tai et~al.(2015)Tai, Xiao, Zhang, Wang et~al.}]{tai2015convolutional}
Tai, C.; Xiao, T.; Zhang, Y.; Wang, X.; et~al. 2015.
\newblock Convolutional neural networks with low-rank regularization.
\newblock \emph{arXiv preprint arXiv:1511.06067}.

\bibitem[{Tucker(1966)}]{tucker1966some}
Tucker, L.~R. 1966.
\newblock Some mathematical notes on three-mode factor analysis.
\newblock \emph{Psychometrika}, 31(3): 279--311.

\bibitem[{Vaswani et~al.(2017)Vaswani, Shazeer, Parmar, Uszkoreit, Jones,
  Gomez, Kaiser, and Polosukhin}]{Attention}
Vaswani, A.; Shazeer, N.; Parmar, N.; Uszkoreit, J.; Jones, L.; Gomez, A.~N.;
  Kaiser, L.~u.; and Polosukhin, I. 2017.
\newblock Attention is All you Need.
\newblock In Guyon, I.; Luxburg, U.~V.; Bengio, S.; Wallach, H.; Fergus, R.;
  Vishwanathan, S.; and Garnett, R., eds., \emph{Advances in Neural Information
  Processing Systems}, volume~30. Curran Associates, Inc.

\bibitem[{Xu et~al.(2018)Xu, Wang, Zhou, Lin, and Xiong}]{xu2018deep}
Xu, Y.; Wang, Y.; Zhou, A.; Lin, W.; and Xiong, H. 2018.
\newblock Deep neural network compression with single and multiple level
  quantization.
\newblock In \emph{Proceedings of the AAAI Conference on Artificial
  Intelligence}, volume~32.

\bibitem[{Zhuang et~al.(2018)Zhuang, Tan, Zhuang, Liu, Guo, Wu, Huang, and
  Zhu}]{zhuang2018discrimination}
Zhuang, Z.; Tan, M.; Zhuang, B.; Liu, J.; Guo, Y.; Wu, Q.; Huang, J.; and Zhu,
  J. 2018.
\newblock Discrimination-aware channel pruning for deep neural networks.
\newblock In \emph{Advances in Neural Information Processing Systems},
  875--886.

\end{thebibliography}

\end{document}